\UseRawInputEncoding
\documentclass[letterpaper, 10 pt, conference]{ieeeconf}
\IEEEoverridecommandlockouts                                                                                
\usepackage{tikz}
\usepackage{textcomp}
\usepackage[bookmarks=false]{hyperref}
\usepackage{lipsum}

\newcommand\copyrighttext{%
  \footnotesize \textcopyright 2021 IEEE. Personal use of this material is permitted.
  Permission from IEEE must be obtained for all other uses, in any current or future 
  media, including reprinting/republishing this material for advertising or promotional 
  purposes, creating new collective works, for resale or redistribution to servers or 
  lists, or reuse of any copyrighted component of this work in other works.  
  }
\newcommand\copyrightnotice{%
\begin{tikzpicture}[remember picture,overlay]
\node[anchor=south,yshift=10pt] at (current page.south) {\fbox{\parbox{\dimexpr\textwidth-\fboxsep-\fboxrule\relax}{\copyrighttext}}};
\end{tikzpicture}%
}

\usepackage{cite}
\usepackage{amsmath,amssymb,amsfonts}
\usepackage{algorithmic}
\usepackage{graphicx}
\usepackage{textcomp}
\usepackage{xcolor}
\usepackage{mdframed}
\usepackage{subfig}
\usepackage{float}
\usepackage{microtype}
\def\BibTeX{{\rm B\kern-.05em{\sc i\kern-.025em b}\kern-.08e m
    T\kern-.1667em\lower.7ex\hbox{E}\kern-.125emX}}

 \newcommand \W{\mathcal{W}}
 \newcommand \OM{\mathcal{\tilde{W}}}
 \newcommand \E{\mathcal{E}}

\begin{document}

\title{CoPEM: Cooperative Perception Error Models for Autonomous Driving}

\author{Andrea Piazzoni$^{1,2}$,  Jim Cherian$^{2}$, Roshan Vijay$^{2}$,\\ 
Lap-Pui Chau$^{3}$, \IEEEmembership{Fellow, IEEE},
Justin Dauwels$^{4}$, \IEEEmembership{Senior Member, IEEE}
 \thanks{
This work was supported in part by the Centre of Excellence for Testing \& Research of AVs - NTU (CETRAN), under the Connected Smart Mobility (COSMO) programme.}
  \thanks{$^{1}$ERI@N, Interdisciplinary Graduate Programme, Nanyang Technological University, Singapore  {\tt\small  andrea006@ntu.edu.sg}}
 \thanks{$^{2}$Centre of Excellence for Testing \& Research of AVs, Nanyang Technological University, Singapore {\tt\small  jcherian@ntu.edu.sg, rvijay@ntu.edu.sg}}
  \thanks{$^{3}$School of Electrical and Electronic Engineering, Nanyang Technological University, Singapore   {\tt\small  elpchau@ntu.edu.sg}}
 \thanks{$^{4}$TU Delft, Dept. of Microelectronics, Fac. EEMCS, Mekelweg 4 2628 CD, Delft.   {\tt\small  j.h.g.dauwels@tudelft.nl}}
  }

\maketitle
\copyrightnotice
\thispagestyle{empty}
\pagestyle{empty}

\begin{abstract}
In this paper, we introduce the notion of Cooperative Perception Error Models (coPEMs) towards achieving an effective and efficient integration of V2X solutions within a virtual test environment.
We focus our analysis on the occlusion problem in the (onboard) perception of Autonomous Vehicles (AV), which can manifest as misdetection errors on the occluded objects.
Cooperative perception (CP) solutions based on Vehicle-to-Everything (V2X) communications aim to avoid such issues by cooperatively leveraging additional points of view for the world around the AV.
This approach usually requires many sensors, mainly cameras and LiDARs, to be deployed simultaneously in the environment either as part of the road infrastructure or on other traffic vehicles. However, implementing a large number of sensor models in a virtual simulation pipeline is often prohibitively computationally expensive. Therefore, in this paper, we rely on extending Perception Error Models (PEMs) to efficiently implement such cooperative perception solutions along with the errors and uncertainties associated with them. We demonstrate the approach by comparing the safety achievable by an AV challenged with a traffic scenario where occlusion is the primary cause of a potential collision.
\end{abstract}

\section{Introduction}
AVs need to perceive the environment around them to drive safely. In a typical configuration, an AV relies only on its onboard Sensing and Perception subsystem (S\&P) to achieve this goal. This solution, while independent, is intrinsically limiting, as it implies a single Point of View (PoV); that of the ego-vehicle itself. This may cause occlusion problems, leading to perception errors, e.g., misdetection, and potential safety hazards. For example, on March 24\textsuperscript{th} 2017, a human driver hit an occluded, oncoming AV at an intersection \cite{germbek2018safe}.

Cooperative Perception (CP), i.e., third-party external perception information communicated to the AV via V2X protocols, provides an avenue to overcome this limitation, as it enables the AV to leverage multiple PoVs. Specifically, various infrastructure-based (V2I) roadside perception units (RSPU) or even other connected vehicles (V2V), that communicate with an Autonomous Vehicle (AV) using V2X through Roadside Units (RSU) and Onboard Units (OBU), may be collaboratively combined with its onboard perception (OBP) output. This study focuses on the effects on AV safety given augmented perception information provided by such third-party sources. 

In recent years, virtual testing has been well-established as an indispensable tool for validating the safety of AVs. However, achieving an effective integration of V2X-assisted CP within a simulation environment remains challenging and computationally expensive, more so when considering high fidelity simulations of multiple infrastructure and AV S\&P instances. The AV under test may itself have many types of complex sensors. Alternatively, one can run virtual tests without  S\&P by relying on the ground truth, which is readily available from the simulation tool. This solution may provide results that do not necessarily correspond to the real world as perception errors are often ignored.

In our previous works \cite{ijcai2020-483}, we have introduced the notion of Perception Error Models (PEM) as a convenient and efficient solution to incorporate perception errors into a simulation pipeline. PEMs directly model such errors and apply them on the ground truth, effectively injecting errors in-the-loop with the benefit of not needing to generate or simulate synthetic sensor data. If simulating just a single high-fidelity AV with a few sensors seems computationally expensive, simulation of V2X CP for large deployment areas, requiring simultaneous simulation of multiple S\&P instances, can quickly become infeasible.
In this paper, we aim to expand the PEM approach by including V2X use cases such as CP and model the additional perception errors and uncertainties involved.

In this paper, we make the following contributions:
\begin{itemize}
    \item We define cooperative PEMs (coPEMs) as the extension of PEMs for V2X-based CP applications.
    \item We show how to model V2X-relevant perception errors, implement them as coPEMs, and integrate them into a simulation pipeline.
    \item We demonstrate the approach by comparing the safety achievable by an AV when challenged by a traffic scenario where occlusion can lead to a potential collision.
\end{itemize}

\begin{figure}[t]
\centering
\includegraphics[width=\columnwidth]{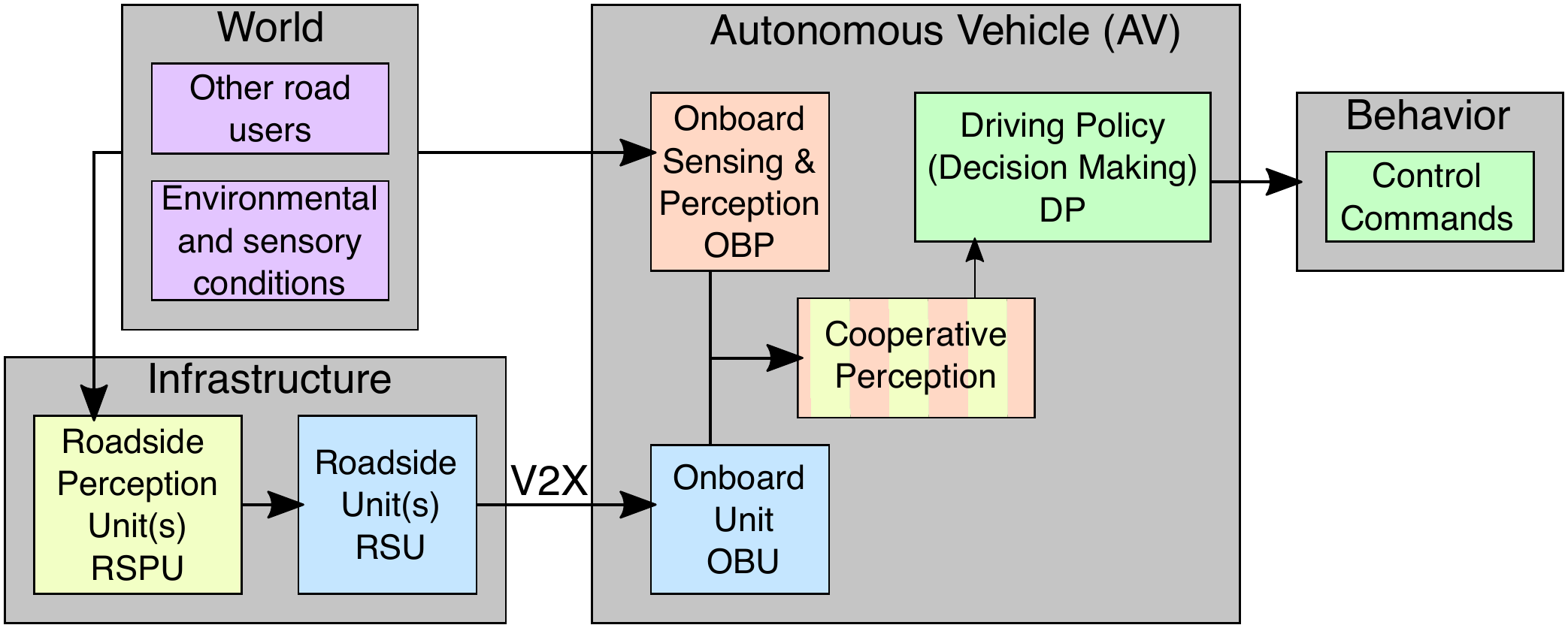}
\caption{Conceptual overview of AV  Perception (both Onboard and Cooperative), in a V2X-based deployment.
}\label{fig:Arch}
\end{figure}

\section{Related works}
The field of V2X communication for road safety-related services has been well studied in the past two decades. The communication aspects and some typical use cases have been standardized by organizations such as SAE International and ETSI. Examples include the SAE J2735 for DSRC message format \cite{standard2020j2735} and the ETSI TR 103 562 \cite{etsi_TR_103_562_2019} describing the CPM message format for Collective Perception Service (CPS). These standards focus on the information that can be exchanged to provide a safer and more efficient use of the transportation network, such as road layout, safety alerts, signal phase and timing, and data on other road users and vehicles. Evidently, the communication component of V2X requires in-depth testing, analysis \cite{schiegg2019Analytical}, and simulation \cite{ZhangSim}, as bandwidth and latency can play a major role in the overall quality of the system \cite{Pilz2021coop}.
In this paper, we focus on the task of V2X-enabled CP, i.e., multiple perception-capable units such as On-board Perception (OBP) and Roadside Perception Units (RSPU) augment their perception via V2X communication \cite{gunter2016Realizing}. 
Identifying the location of RSPUs and their sensors is a major problem in this context, as their optimal placement can drastically improve safety and reduce cost (e.g., procurement, installation, and maintenance). Many studies aim to optimize coverage \cite{geissler2019optimized,roshan2021Optimal}.
To achieve this, some authors consider probabilistic coverage, introducing probabilistic sensor models  \cite{akbarzadeh2012probabilistic,akbarzadeh2014efficient,argany2018optimization}.

Recent works are also studying synthetic data generation for CP  \cite{arnold2021fastreg,li2022v2x}. For example, the open-source CARLA simulator \cite{Dosovitskiy17} is used to simulate the traffic environment and generate synthetic data.

Instead of RSU/RSPU placement or communication protocols, in this study, we look at V2X from the perspective of AV safety. 
Virtual simulation is a safe and convenient approach \cite{young2014} for validation of AV safety. 
Shan et al. demonstrated CP in a simulation environment relying on synthetic data \cite{shan2020demonstrations}. This approach is limited by the bottleneck of computational cost in the case of more than a few RSPUs or OBUs.
Thus, we propose a framework for AV safety testing that includes perception errors and supports V2X solutions, i.e., an indefinite number of OBPs, OBUs, RSUs, and RSPUs.

\section{Approach}
In this section, we introduce Cooperative PEM (coPEM). First, we provide a summary background on the notation we employ (see \autoref*{fig:Arch}.
We abstract an AV as a composition of a Sensing and Perception module (S\&P) and the Driving Policy (DP; the decision-making module)  \cite{ijcai2020-483}. The S\&P has the task of observing the surrounding objects $o_j$, included in the surrounding world $\W$, to generate the perceived world $\tilde \W$, i.e., the object list describing the perceived surrounding obstacles $\tilde o_j$.
The object list $\tilde \W$ is then analyzed by DP to determine the AV response.

In the case of CP, let $N$ be the number of the third-party Perception Units (PUs; RSPUs and other OBPUs), each identified by $n \in \{1,\ldots, N\}$, with ego-vehicle index 0. 
We assume that each PU$_n$ perceive the surrounding world with its S\&P system: $\tilde \W_n = \text{S\&P}_n(\W)$, following the same abstraction of AVs.
The $N+1$ object list $\tilde \W_n$, are then fused together: $\tilde \W = f(\tilde \W_n)$, where $f$ implements the CP fusion task.
We assume the fusion is at feature-level \cite{Kaempchen2005feature}, which is suitable for V2X solutions as the communication can be performed at the objects level.

\subsection{Individual PEM}
In our previous work \cite{ijcai2020-483}, we defined Perception Error Model (PEM) for AVs.
A PEM is a model that receives the ground truth world $\W$  and returns the perceived world $\OM$. 
Since virtual environments provide direct access to the ground truth world $\W$, a PEM can be integrated into a simulation pipeline as a flexible and efficient surrogate of the S\&P.
This approach removes the need for synthetic signals and introduces explicit perception errors at the object level. 
For example, a PEM can alter the perception of an object's position, or remove it from the object list, or more.

We define a PEM as an approximation of the combined function of sensing (S) and a perception (P) subsystems:
    \begin{equation}\label{eq:pemFunction}
    \text{PEM}(\W) \approx \text{S\&P}(\W) = \OM = \W + \E,
    \end{equation}
where $\E$ is the perception error.
We do not prescribe any preferred approach for a PEM implementation, but we opt for probabilistic models for a demonstration.
In this article, we implement PEM$_n$ as a set of $|C|$ couples:

\begin{equation}\label{eq:pem}
    \text{PEM} = \{(d_c, p_c), c \in C\}.
\end{equation}
where:
\begin{itemize}
    \item $C$ is the set of conditions that alter the perception quality.
    \item $d_c$ is the detection model, i.e., the model that    determines if an object is detected. 
    \item $p_c$ describes the parameter error $\varepsilon$ distribution, where $\varepsilon_j = \tilde o_j - o_j$, i.e. the difference between perception output and ground truth.
\end{itemize}
Generally, we include positional aspects (i.e., Field of View, operational range) in $C$.
This notation affords a flexible implementation, and a single PEM may include different probabilistic models given the condition $c$. 
For example, a couple $(d,p)$ can be dedicated to model the frontal area of the PU, with high detection rates and low parameter errors. Conversely, another $(d,p)$ associated with a further away area may have lower detection rates and higher parameter errors.

A PEM$_n$ generates $\tilde \W_n$ by processing each object $o_j\in \W$. First, we select the $(d,p)$ to employ, by determining the current condition $c$, e.g., the relative position.
Then, we compute if the object is detected via $d(o_j)$, i.e. if it is included in the $\tilde \W$. 
Lastly, we sample the parameter error $\varepsilon_j$ according to the distribution $p(\varepsilon|o_j)$.

\subsection{CoPEMs}
The previous definition of PEM does not consider external sources of information, i.e., CP elements:
\begin{itemize}
    \item Heterogeneous Units: The S\&P deployed in an AV is not self-contained anymore, and the DP does not directly consume its output $\tilde \W$. 
    \item Dynamic Relative Position: Third-party RSU and OBU, static (V2I) or dynamic (V2V), have a dynamic relative position w.r.t. the ego-vehicle.
    \item Trust, Reliability, Latencies, and Uncertainties: Not all units have the same reliability. For example, static units potentially have lower uncertainties, while surrounding vehicles can be less precise or even malicious and not trustworthy \cite{ambrosin2019Design}. In this study, we limit the analysis to the uncertainties and do not consider security issues.
\end{itemize}

\begin{figure}[t]
\centering
\includegraphics[width=\columnwidth]{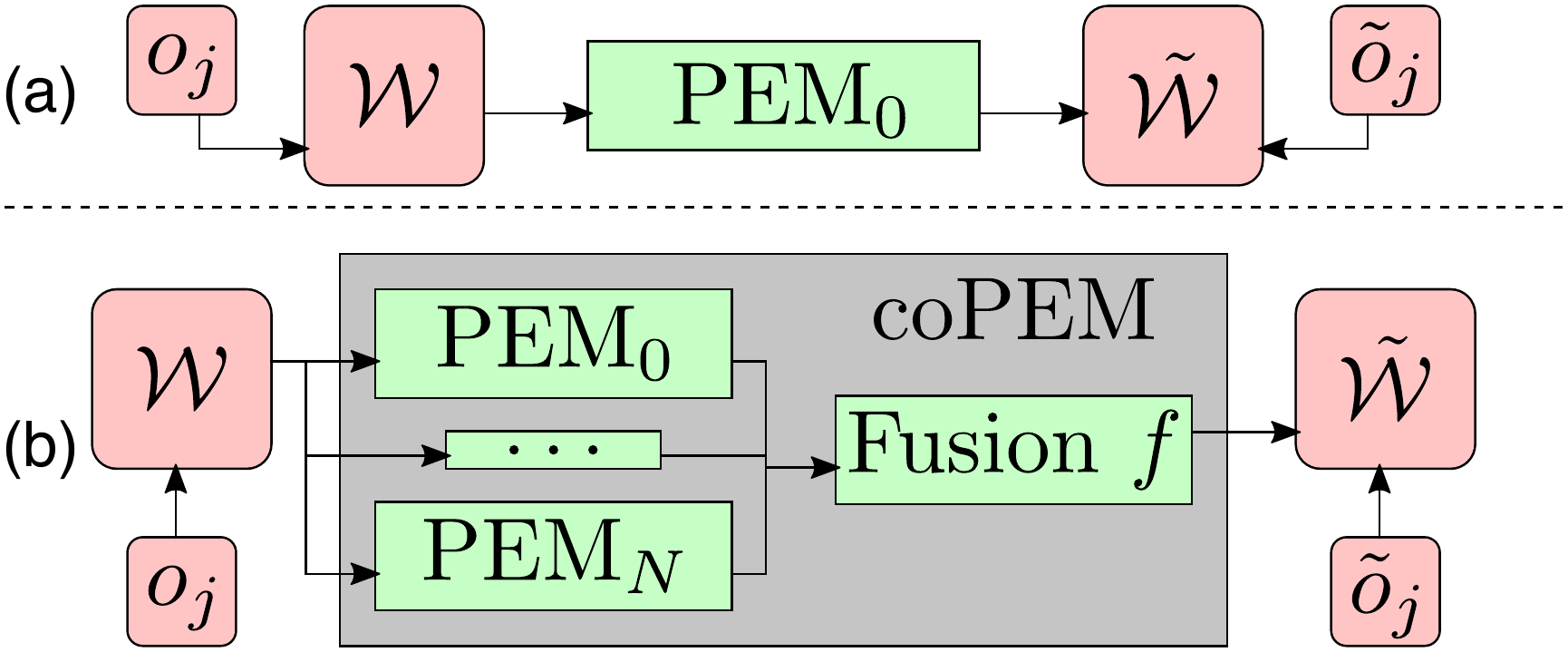}
\caption{Illustrations of (a) a PEM and (b) a coPEM as a collection of PEMs.
Both the PEM and the coPEM receives $\W$ in input and return $\tilde \W$ in output, in fact, they have the same interface.
In the coPEM, each PEM$_n$ generates a $\tilde \W_n$, i.e. the perception output of each Perception Unit, and through a fusion phase (i.e., CP) are combined into the final $\tilde \W$.}\label{fig:copem}
\end{figure}

However, a PEM can model each independent PU since we can assume that each of them has an S\&P system.  
Hence, we define coPEM as a CP fusion function $f$ and a set of PEMs representing the deployed $N+1$ perception units (i.e., the $N$ third-party PUs and the ego vehicle's OBP).
    \begin{equation}\label{eq:copemFunction}
    \text{coPEM} = (f;\{ (\mathbf{X}_n,\text{PEM}_n), 0\leq n\leq N\}). 
    \end{equation}
Where $\mathbf{X}_n$ indicates the pose of the unit.
Given the dynamic location of the AV, we need to keep the model flexible and be able to compute the relative positions of all the PU involved in each step. Hence, in each step, $\mathbf{X}_n$ can be updated to keep the models aligned.
In \autoref*{fig:copem}, we outline the differences and similarities between the structures of PEMs and coPEMs.

\newcommand{\rulesep}{\unskip\ \vrule height 32mm width .01mm}

\begin{figure*}[t]
\centering
\subfloat[][]{\includegraphics[width=.95\columnwidth]{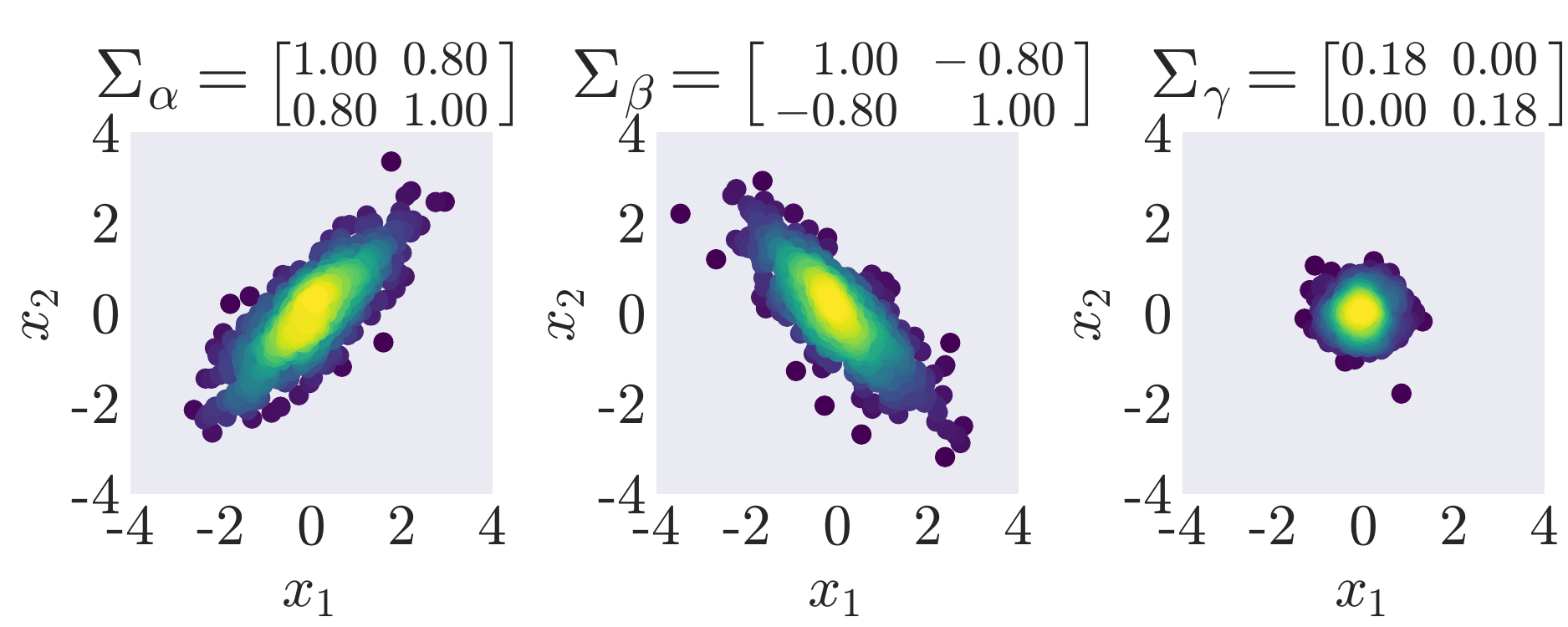}}
\hspace{2mm}
\rulesep
\hspace{2mm}
\subfloat[][]{\includegraphics[width=.95\columnwidth]{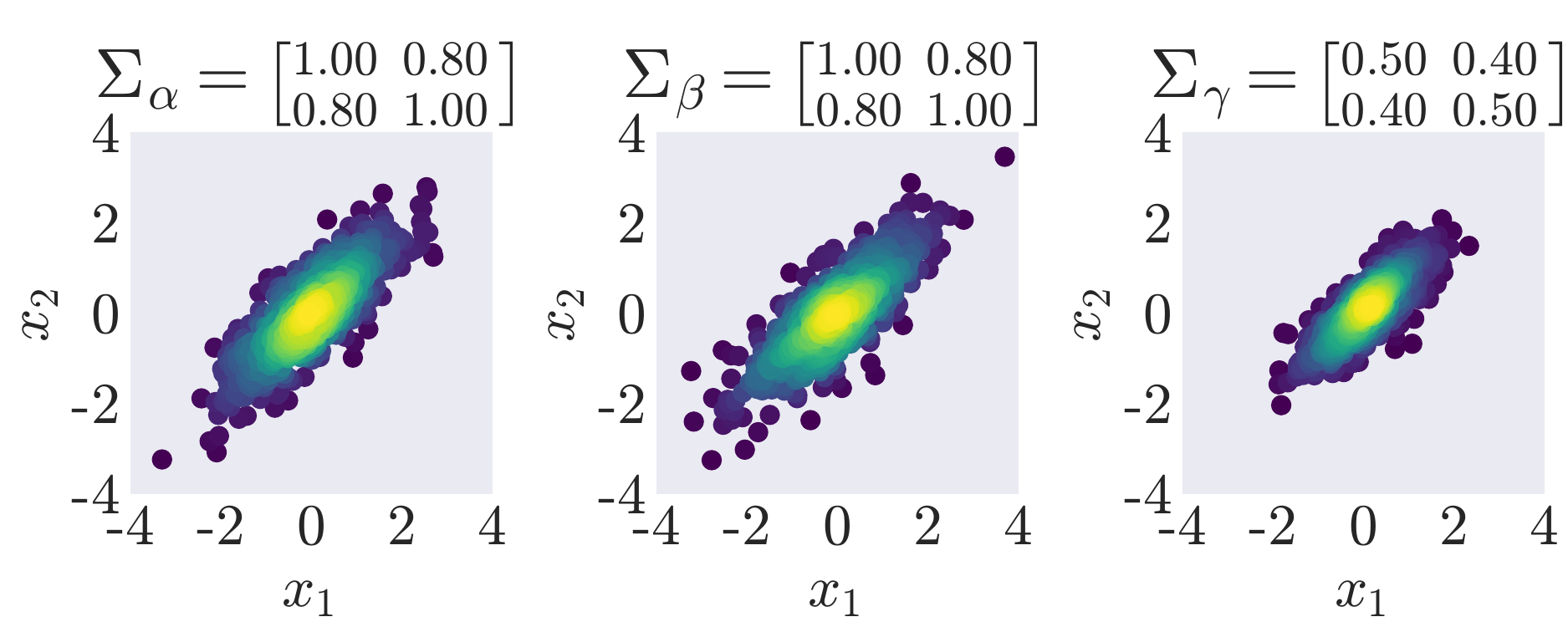}}
\caption{Sampling from 2 Gaussian distributions ($\alpha$,$\beta$) and from their combination obtained by applying \autoref*{eq:var} ($\gamma$).
(a) The distributions $\alpha$ and $\beta$ have different covariance matrices.
We observe that $\Sigma_\gamma$ has lower values, and its shape roughly resembles the intersection.
In (b), instead, the distributions  $\alpha$ and $\beta$ have equal covariance matrices. Hence, their combination $\Sigma_\gamma$ has the same shape and slightly smaller values. This approach is the multivariate equivalent of averaging multiple measures to increase accuracy.}
\label{fig:samecross}
\end{figure*}

\subsection{Ideal Fusion Model}
The newly introduced component to PEM is the fusion model $f$. Once the $\tilde o_{n}$ are generated, we outline two approaches for the fusion step:
\begin{itemize}
    \item Perception Fusion Algorithm $f$: Since the coPEM receives $\W$, we can apply each PEM$_i$ to generate all the $\tilde \W_i$. Then, we can integrate an actual perception fusion algorithm $f$ to generate the consolidated $\tilde \W$. This approach is preferable when the focus of the virtual testing is this fusion algorithm itself.
    \item Ideal Fusion Model: Alternatively, we can model an optimal function $f$ and directly generate $\tilde \W$. This option affords a simpler simulation as we can study other aspect of CP while discounting additional errors due to an not optimal $f$.  
\end{itemize}

The first solution avoids any assumption, as an actual fusion algorithm is integrated into the pipeline. 
We propose an Ideal Fusion Model by assuming an optimal algorithm:
\begin{itemize}
    \item Matching: The first step of merging $M$ units is to find the matching of each object $\tilde o_j$ within multiple $\tilde \W_m$.
    An optimal $f$ would perform this task without any error, avoiding false positives. 
    \item Parameter Errors: Despite perfect matching, the objects $\tilde o_{jm}$ within each object list $\OM_m$ will be affected by an error $\varepsilon_{jm}$, as each PU is affected by errors. The optimal $f$ would minimize the resulting error $\varepsilon_{j}$, i.e., the error affecting $\tilde o_j$ after the fusion.
\end{itemize}
By leveraging the virtual environment and PEMs, we can implement an optimal $f$. We can exploit the underlying ground truth $\W$ to craft an optimal matching. Moreover, PEMs provide a probabilistic distribution for the parameter errors, which we can use to weigh the reliability of the PUs.

As a first step, we must determine the $M_j$ units are actively involved in each object detection. Thus, each PEM$_n$ should determine if the unit $n$ detects an object $o_j$ and participates in its parameter estimation.
To combine parameter errors, we can formulate a measure combination problem.
Each unit $m \in \{1,\ldots, M_j\}$ participating in the detection of an object $o_j$ provides a vector of measurements for each physical property (i.e., the object parameters) with an error $\varepsilon_{jn}$. We model the distribution of each error $\varepsilon_{jn}$ in each PEM$_n$.
Thus, we need to determine the posterior distribution of $p(\varepsilon_j|o_j)$.

Bayesian inference can solve this problem, and the joint distribution can be written as:
\begin{equation}
    p(\varepsilon_{0}|o_j) \ldots p(\varepsilon_{M}|o_j)p(o_j).
\end{equation}
where $p(\varepsilon_{m}|o_j)$ is the distribution of the error $\varepsilon_m$  given the object $o_j$. Instead, $p(o_j)$ is the prior of the model, which describes the dynamics of the object.

In the case of Gaussian distributions, the inverse-variance weighted average minimizes the variance of the average \cite{shalizi2013advanced}. This way, more precise units (i.e., lower variance) have more influence on the outcome, and we imply an optimal combination of the units.
We can write the V2X outcome distribution of the error  $\varepsilon$ as:w
\begin{equation} \label{eq:post}
p(\varepsilon_j|o_j)  \sim \mathcal{N}(\mu ,\Sigma),
\end{equation}
Where mean and covariance are calculated as:
\begin{equation}\label{eq:mean}
   \mu = \left(\sum_n{\left(\Sigma_n^{o_j}\right)^{-1}}\right)^{-1}\sum_n{\left(\Sigma_n^{o_j}\right)^{-1}\mu_n^{o_j}}.
\end{equation}
\begin{equation}\label{eq:var}
   \Sigma = \left(\sum_n{\left(\Sigma_n^{o_j}\right)^{-1}}\right)^{-1}.
\end{equation}
We note that, in Equations \ref*{eq:post}, \ref*{eq:mean}, and \ref*{eq:var}, parameters $\mu_n^{o_j}$ and $\Sigma_n^{o_j}$ are conditioned to $o_j$ via $c$ (see \autoref*{eq:pem}).

This approach implies that each parameter is modeled in absolute values. 
We avoid relative values such as distance and azimuth from ego-vehicle. Instead, we model absolute coordinates (e.g., northing and easting).  
With these assumptions and procedures, we can use $p(\varepsilon|o_j)$ to sample the perception error.
In \autoref*{fig:samecross}, we show two examples with Gaussian distributions.

\subsection{V2X Latency}
As previously mentioned, latency can be a significant bottleneck in V2X solutions \cite{Pilz2021coop}.
Although it is not an issue in a virtual environment, latency can be specifically modeled and introduced to improve realism in virtual V2X testing.
PEMs and coPEMs provide a straightforward solution to inject latency.
When generating $\tilde \W^t$ at time $t$, instead of applying the model to the current world state $\W^t$, we can instead consider a previous $\W^{t-l}$ where $l$ is the overall latency. The parameter $l$ indicates the time difference between when an event happens and when the AV DP receives the information.

This solution effectively implies a slower reaction time induced by the communication delay.
We do not consider, in this study, heterogeneous delays, as we assume a single $\tilde \W$ is consolidated each time frame.

\section{Experiments}
We conduct a series of experiments under specific configurations to demonstrate our proposed approach.
With these experiments, we aim to show the impact of occlusion on safety and how cooperative perception may address it.
Each experiment is a set of simulations run identified by two elements: a perception configuration and a test case (i.e., a parametrized scenario).
We implement our experiment in the ViSTA framework \cite{piazzoni2021vista}, which interfaces with SVL simulator \cite{LGSVL} and Apollo \cite{Fan2018} as the AV under test. We choose SVL since it facilitates the co-simulation by providing connection bridges toward independent AV stacks.

\subsection{Perception Configurations}
We aim to compare the different behavior induced by the varying perception quality, 
Thus, we consider three perception configurations, with additional variations for a total of 6 configurations (see \autoref*{table:setups}):
\begin{itemize}
\item GT: The simulation is based on ground truth, and occlusion is not a factor. The AV can detect anything in the environment with no delay.
\item PEM: We replace the AV's S\&P with a PEM. Occlusion has a significant role, and the detection rate of the model $d(o_j)$ is equal to the portion of the surface visible from the AV continuously from 0\% for a completely occluded object, to 100\% for a completely visible one. Moreover, we model an error in object location with a multivariate Gaussian distribution over northing and easting with variance of 1 m in both directions.
\item coPEM: The AV's PEM is the same as the previous configuration. However, it is supported by $N=20$ external V2X units, each modeled by its PEM. We tested four variations with increasing latency (0.5s increments).  
\end{itemize}
With this set of configurations, we can observe the effects occlusion (PEMs) and latency (coPEMs) have on safety, and we can consider GT as the baseline.

\begin{table}[t]
\centering
\caption{List of Perception Configurations.}\label{table:setups}
 \begin{tabular}{l | l } 
 ID & Sensor configuration \\
 \hline
 GT & Ground Truth based.\\ 
 PEM & A PEM injects perception errors.\\
 coPEM:0s & coPEM with 0s latency.\\
 coPEM:0.5s & coPEM with 0.5s latency\\
 coPEM:1s & coPEM with 1s latency.\\
 coPEM:1.5s & coPEM with 1.5 latency\\
\end{tabular}
\end{table}

\subsection{Test Cases and Preliminary Exploration}
The second element of each experiment is the test case.
Each test case is defined by an AV starting point, a destination, and scripted actors with pre-defined behavior. 
In our previous work presented at the IEEE Autonomous Driving AI Test Challenge \cite{piazzoni2021vista}, we have defined and implemented a comprehensive set of Test Cases for testing AVs.
\begin{figure}[t]
\centering
\includegraphics[width=\columnwidth]{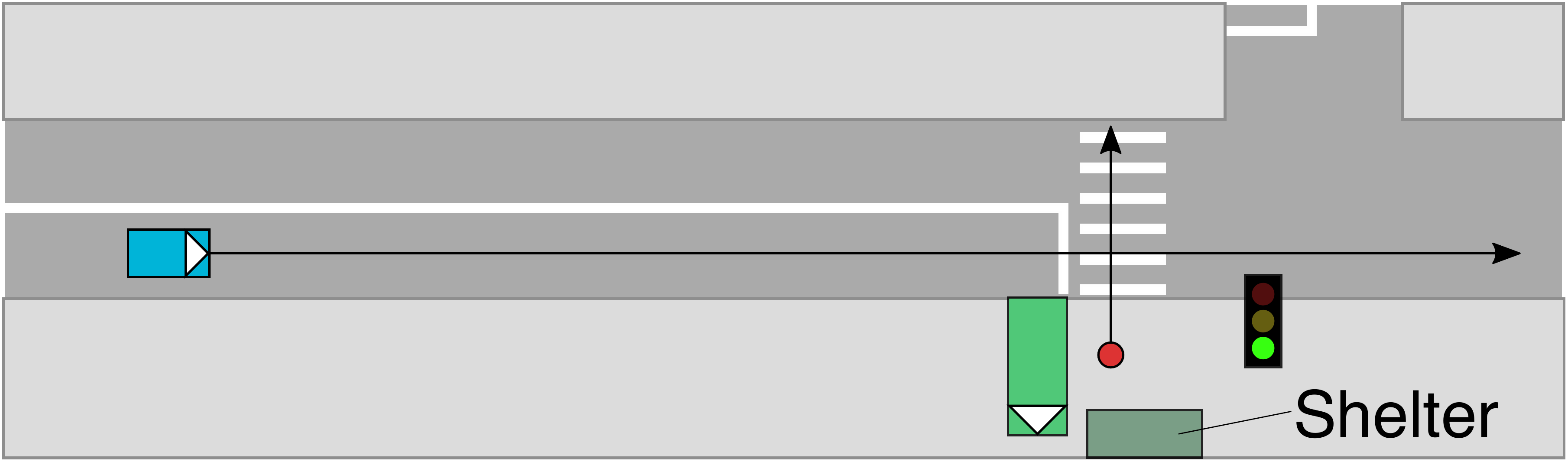}
\caption{Illustration of the Test Case. The AV (blue) approaches a pedestrian (red) occluded by a parked truck (green).}
\label{fig:scenario}
\end{figure}

As the first step, we conducted a preliminary exploration phase.
We tested each of the $\sim30$ scenarios with various perception configurations to identify interesting variations in AV behavior due to occlusion issues.

For a deeper analysis, we selected a scenario inspired by TC\_016.d from \cite{piazzoni2021vista}.
This test case was problematic to implement under the existing setup \cite{piazzoni2021vista}, i.e., using ground truth, as occlusion was not modeled.  
In this scenario, the AV approaches a signalized intersection on an urban road.
A truck parked on the sidewalk occludes a pedestrian near a bus shelter, as illustrated in \autoref*{fig:scenario}.
The pedestrian begins to cross once the AV approaches, even if the AV has the right of way as indicated by the green light.

We considered slight variations of this scenario by changing the AV path, location (i.e., introducing specific road features that can induce occlusion, such as a road bend), or tuning other parameters.
We found that Apollo generally employs a conservative driving style and can handle most road features or maneuvers.
Thus, we focused on parameter variations, employing GT and PEM configurations to achieve a meaningful test case.
We scripted the test case to have the following properties:
\begin{itemize}
    \item Plausible: The parameters applied (e.g., pedestrian walking speed) should be in a plausible range or, at most, incorporate edge cases.
    \item Fair: The test case should allow a safe outcome, and we expect a high success rate under GT configuration.
    \item Challenging: The test case should not be trivial, and we expect failures in cases of significant perception errors.
\end{itemize}

We found relevant parameters as follows:
the AV starts far enough to reach a cruising speed of $\sim10m/s$, and the pedestrian starts walking at $1m/s$ when the AV is within a $20m$ range.
These settings lead to a potential collision $\sim 2s$ after the pedestrian starts walking. Given the occlusion and V2X latency, however, the AV may not be able to detect the danger instantaneously, resulting in a delayed reaction.
We note that Apollo does not perform any evasive maneuver. Thus, the collision may only be avoided by braking.

Moreover, we experimented with RSPU positioning, and simplified the experiment set by considering only optimal placements (refer \cite{roshan2021Optimal} for details) that actually help overcome the occlusion.
Without this step, there could be redundant RSPU placements and coPEM configurations that may result in the same or similar AV behavior as with the regular PEM configurations (non-cooperative perception).

\newcommand\sizescatter{.22\textwidth}

\newcommand\myhspace{\hspace{2mm}}
\begin{figure}[t]
     \centering

       \includegraphics[width=\columnwidth]{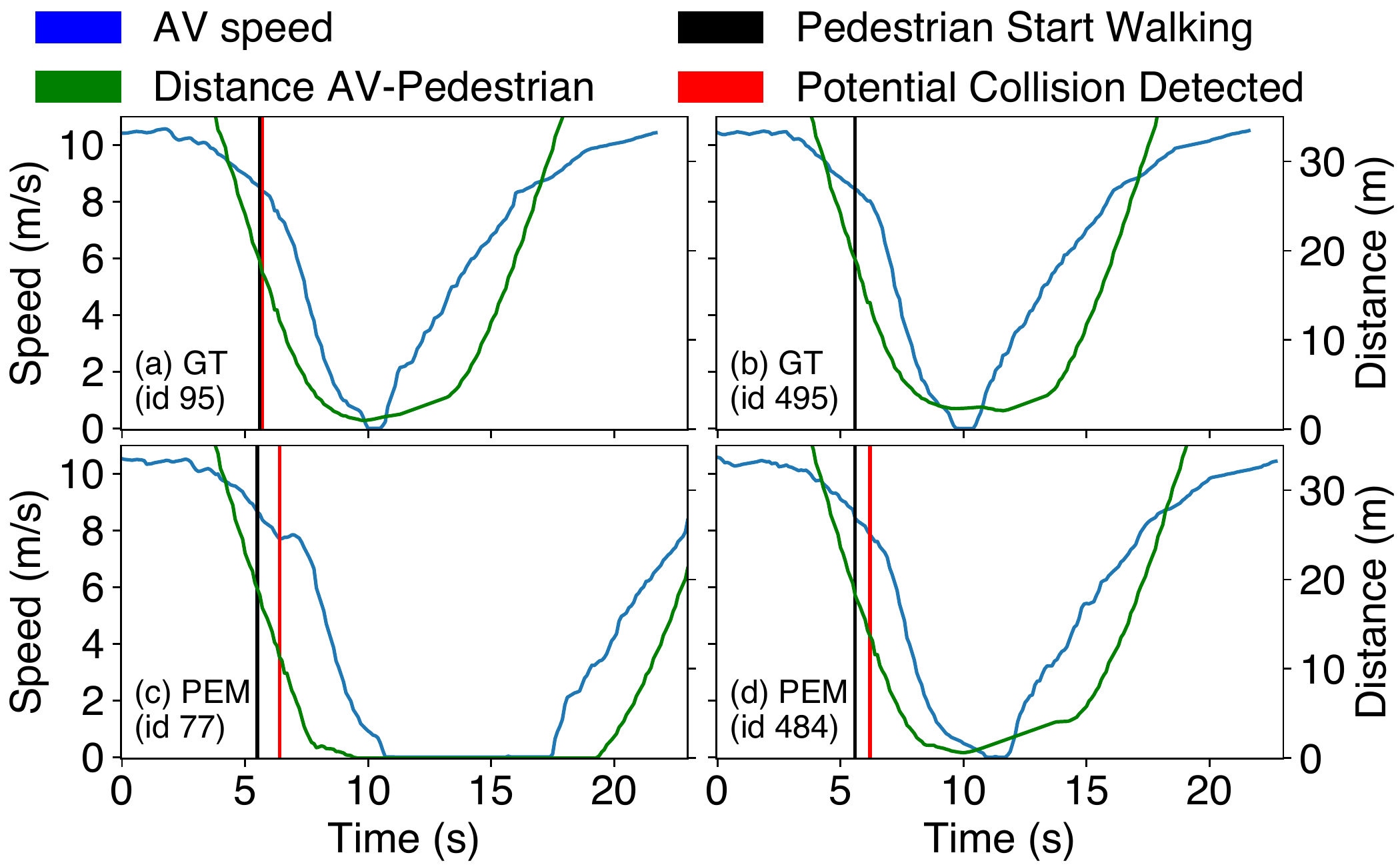}
      \caption{Illustration of the relevant parameters and events of four example simulation runs. 
      }
     \label{fig:runPlots}
\end{figure}

\begin{figure}[t]
     \centering
     \includegraphics[width=\columnwidth]{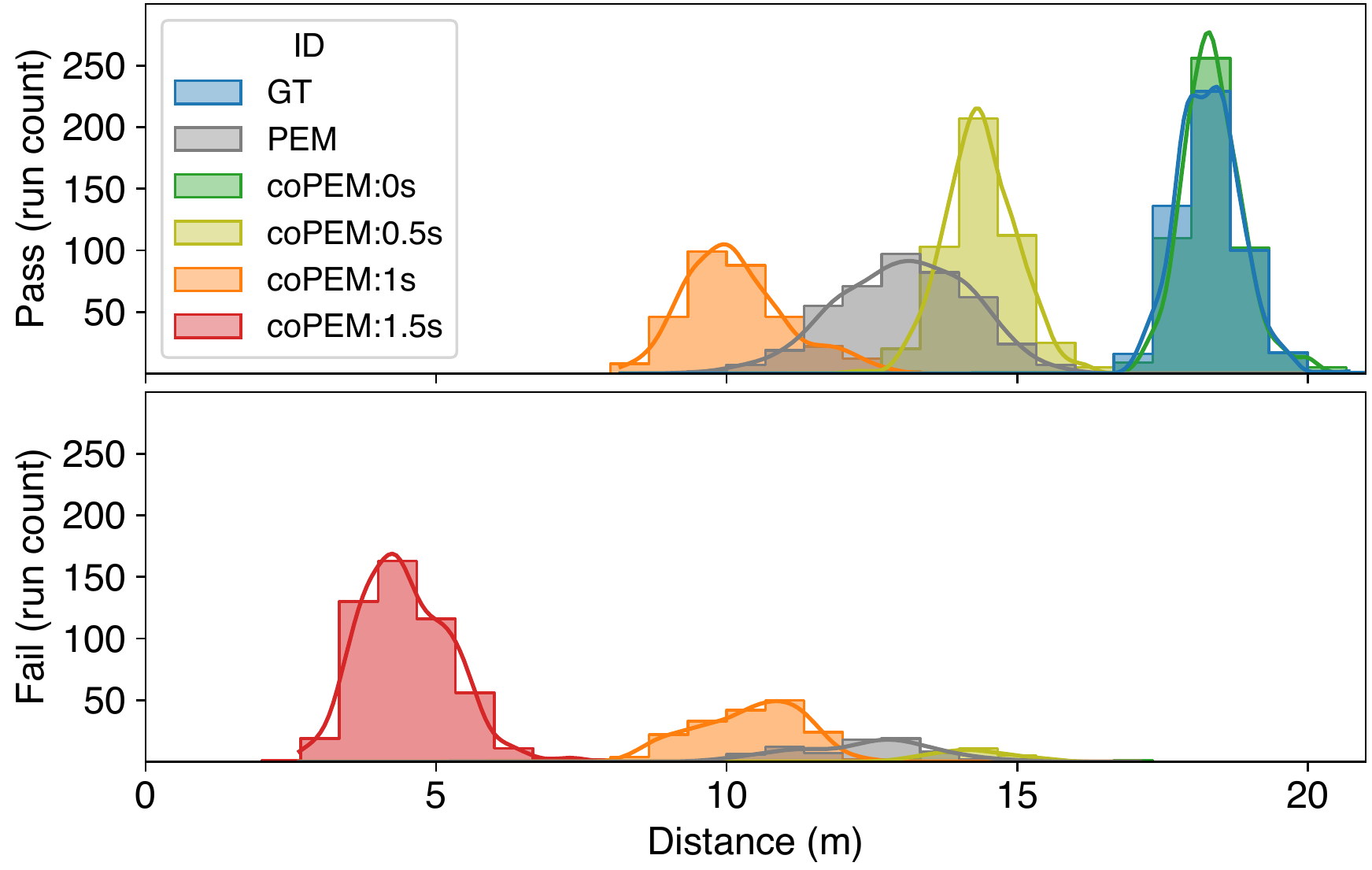}
 \caption{Summarized results from 3000 runs (500 for each Perception Configuration). For each run, we consider a data-point, i.e.,  the distance between the AV and the pedestrian at the moment of detection of the potential collision.}
     \label{fig:warning}
\end{figure}

\subsection{Results}
Due to the page limit, we do not report results for all the explored scenarios and parameter variations.
We simulated 500 runs for each perception configuration, i.e., 3000 runs.

During the test case selection and configuration phase, we employed plots of the logged simulation runs \autoref*{fig:runPlots}. These plots facilitate the exploration and parametrization of the test cases.
In each plot, we depict the run dynamics, and we report the most relevant parameters and events.
We can see how the potential collision is immediately detected under the GT configuration, and the AV can brake accordingly.
However, under PEM configuration, the delay of the detection may (\autoref*{fig:runPlots}c) or may not (\autoref*{fig:runPlots}d) lead to a collision.

We categorize the outcome of the runs as Pass or Fail: we consider a run as successful if the AV reaches its destination and maintains a distance $>0.5m$ from all actors (vehicles and pedestrians). 
In \autoref*{fig:warning}, we summarize all 3000 runs, Pass and Fail, for each perception configuration. 
We note that the pedestrian was scripted to start walking when the distance is $<20m$, but latency and occlusion can delay detection. In fact, we can see how the distance between the AV and the pedestrian decreases inversely to the latency introduced in the model. 
Simulation results clearly show how occlusion may lead to collisions. Moreover, we observe how CP can help to overcome the problem if the latency is kept low. Higher latency values, however, make the approach detrimental. Therefore, one may infer from this study that the adoption of low-latency communication schemes is essential to a successful deployment of V2X-based CP.

\begin{table}[t]
\centering
\caption{Success Rate under each configuration.}\label{tab:sr}
\label{table:results}
 \begin{tabular}{l|rr||l|rr} 
 ID & \multicolumn{2}{l||}{Success Rate}&ID & \multicolumn{2}{l}{Success Rate}\\\hline
 GT & 500/500& 100\% &
 coPEM:0.5s& 474/500& 94.8\% \\
 PEM & 426/500 &85.2\% &
 coPEM:1s  & 323/500& 64.6\%\\
 coPEM:0s  & 499/500& 99.8\% &
 coPEM:1.5s & 0/500& 0.0\%\\
 \end{tabular}
\end{table}

\subsection{Discussion}
CoPEMs are designed to be seamlessly integrated into a simulation pipeline. 
They provide an avenue for testing CP fusion algorithms, their effects on the AV perception, and thus the resulting behavior and safety.
CoPEM limitations are mainly two: they require a PEM for each PU (i.e., a statistical model of the perception errors) and do not consider communication issues other than latency.

Moreover, we introduce an Ideal Fusion Model, which determines the optimal outcome of any CP approach while considering perception errors and uncertainties. However, this algorithm assumes normally distributed errors on the object parameters and it only applies in virtual environments since it requires access to the ground truth for the matching step. Nevertheless, it facilitates the testing other aspects of CP, e.g., AV safety, RSPU location, or OBPU critical mass in traffic.

\section{Conclusion}
In this paper, we gauged the effects of misdetection errors on AV behavior, and we analyzed how cooperative perception (CP) can improve safety. 
We conducted a series of simulation experiments by testing multiple perception configurations.
These configurations included ground truth, PEMs, and coPEMs (PEMs extended for CP).
These tools provide a convenient way to tune perception performance in simulation.
Moreover, since they abstract the sensor technology, this solution scales well with an increasing number of sensors to support CP applications.
We experimented with a rich set of scenarios to identify relevant conditions and parameters to finalize concrete test cases. We conducted the tests in a popular virtual simulation platform (SVL), and we deployed a  standard open-source AV software stack (Baidu Apollo) to generate the AV behavior.
Simulation results show the effect of perception errors (occlusion and latency) on AV safety in the selected test cases.
In the future, we plan to extend the proposed approach to support heterogeneous latencies involved in CP, and integrate it with other simulators.
Finally, we also aim to deploy CoPEMs towards testing CP with physical hardware in the loop testing configurations, more complex scenarios, and on-road AV trials. 

\bibliographystyle{./bibliography/IEEEtran}
\bibliography{./bibliography/bib}

\end{document}